# $L_1$-norm Error Function Robustness and Outlier Regularization

**Chris Ding**[1] and **Bo Jiang**[2]*

[a]CSE Department, University of Texas at Arlington, Arlington, TX 76019, USA
[b]School of Computer Science and Technology, Anhui University, Hefei, China
chqding@uta.edu, jiangbo@ahu.edu.cn

**Abstract**

In many real-world applications, data come with corruptions, large errors or outliers. One popular approach is to use $L_1$-norm function. However, the robustness of $L_1$-norm function is not well understood so far. In this paper, we present a new outlier regularization framework to understand and analyze the robustness of $L_1$-norm function. There are two main features for the proposed outlier regularization. (1) A key property of outlier regularization is that how far an outlier lies away from its theoretically predicted value does not affect the final regularization and analysis results. (2) Another important feature of outlier regularization is that it has an equivalent continuous representation that closely relates to $L_1$ function. This provides a new way to understand and analyze the robustness of $L_1$ function. We apply our outlier regularization framework to PCA and propose an outlier regularized PCA (ORPCA) model. Comparing to the trace-norm based robust PCA, ORPCA has several benefits: (1) It does not suffer singular value suppression. (2) It can retain small high rank components which help retain fine details of data. (3) ORPCA can be computed more efficiently.

## 1 Introduction

In many real-world applications, data often come with noise. The most common are Gaussian type noises, where the magnitude of noise are usually small (the chances for them to deviate from the mean by more than 3 standard deviations are only 0.3%). Most machine learning algorithms use error-squared cost functions that amount to maximize the log-likelihood of the data distribution [Duda *et al.*, 2001]. In many situations data come with gross, large, significant errors, due to different reasons, for example images are corrupted, data entry/component are missing, errors due to human recording or machine malfunction, and environments where noises are simply non-Gaussian. For these data, the usual error-squared cost functions are no longer appropriate [Kwak, 2008; Wright *et al.*, 2009; Liu *et al.*, 2010]

*Corresponding author

Recently, a popular approach to deal with data which have large/gross errors or outliers is to use the $L_1$ norm, i.e., the sum of absolute values of the errors, instead of squares of errors [Baccini *et al.*, 1996; Wang *et al.*, 2012; Ke and Kanade, 2005; Kwak, 2008; Wright *et al.*, 2009; Peng *et al.*, 2010; Wagner *et al.*, 2012; Lu *et al.*, 2013]. The most intuitive motivation is that for outliers, the errors are large and thus the squared error are even larger and thus dominate the cost function. Using absolute values in cost function diminishes the undue influence of those outliers and thus makes the learning more robust or stable. This understanding is only partially correct. Although the large errors due to outliers are not squared in $L_1$, they are still *large* and thus one would expect they would influence the cost function. However, careful observations show that the influence of outliers in $L_1$ formulations are small. Thus, although $L_1$ function is widely used for its robustness, to the best of our knowledge, the precise nature of this robustness is not well understood. Other approaches also use $L_{2,1}$-norm function on vector data to make the learning more robust to the outliers [Ding *et al.*, 2006; Liu *et al.*, 2010].

In this paper, we study the $L_1$-norm robustness from a new framework of outlier regularization. Toward this goal, we propose an outlier-regularization function. This function detects whether a data point is an outlier and, if it is, minimizes its influence by pulling it back towards a threshold line. This process of pulling it back is equivalent to regularizing it. There are two main aspects of the proposed outlier regularization. (1) One property of the proposed outlier regularization is that as long as a data point is an outlier, how far away it lies from the theoretical predicted value (outlyingness of the outlier) does not affect the final results. This is because once an outlier is regularized, its influence to the theoretical prediction is the same as non-outliers. This agrees with the results of $L_1$ function. (2) Another important feature of outlier regularization is that it has an equivalent continuous representation that closely relates to $L_1$ function. This provides a new way to understand and analyze the robustness of $L_1$ function.

As an application, we apply our regularization function to data reconstruction using principal component analysis (PCA) and propose a new outlier-regularized PCA (ORPCA) model. The small tolerance limit of ORPCA model leads to pure $L_1$-PCA [Ke and Kanade, 2005]. We also compare our ORPCA and with trace norm based robust PCA [Wright *et al.*, 2009; Chandrasekaran *et al.*, 2009] and perform experiments to validate our approach.

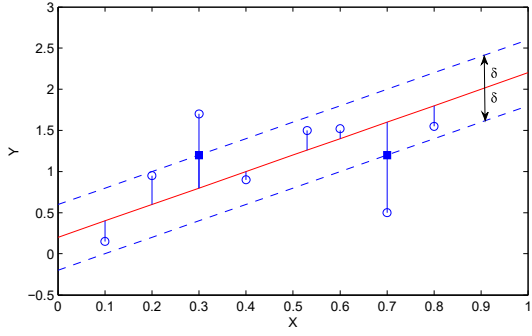

Figure 1: Illustration of outlier regularization. Data points are denoted as circles. Center solid line is the theoretical prediction. Two outliers are regularized: they are pull back to the threshold (dashed) lines.

## 2 Outlier-Regularization Function and Continuous Representation

In this section, we first propose our regularization function and then provide a continuous representation for it.

### 2.1 Outlier-regularization function

Our outlier regularization function is motivated by considering fitting data using linear function. The input data is $(x_i, y_i), i = 1 \cdots n$, as shown in Figure 1. Most data points are associated with the usual small Gaussian random noises. However, two data points are outliers, i.e., the measured $y$ values are significantly distorted or corrupted from their correct values. The correct values are theoretical predictions or the fitted function values $\{f_i\}$. We define a data point as an outlier if the difference between the measured signal $y_i$ and the theoretical prediction $f_i$ is bigger than the tolerance $\delta$. The two dashed lines indicate the tolerance limits in Figure 1.

Now we wish to correct these outliers. One intuitive and effective way is to move the outliers towards the prediction line, but keep them at the boundary (tolerance limit). This is done by the following **outlier regularization function**,

$$\tilde{y}_i = \begin{cases} y_i & \text{if } |y_i - f_i| \leq \delta \\ f_i + \delta \text{sign}(y_i - f_i) & \text{if } |y_i - f_i| > \delta \end{cases} \quad (1)$$

where $\delta > 0$ is a tolerance parameter, $y_i, \tilde{y}_i, f_i$ are scalars.

After this outlier regularization, the outliers no longer influence the regression prediction significantly, because (1) the residual (difference between corrected signals and prediction) are much smaller, and (2) the fact that the number of outliers are in general much less than those non-outlier data points. We will show in the following section that this outlier regularization is closely related to $L_1$-norm function which has been widely used in many applications.

**Comparison with related works.** Note that, as statistic robust methods, Huber loss function [Huber, 1964] and random sample consensus (RANSAC) [Fischler and Bolles, 1981] have been widely used. Different from these models: (1) Our outlier regularization function Eq.(1) provides a way to identify and **regularize/correct** the outliers using theoretical prediction, i.e., the outliers are pulled back to threshold line, as shown in Eq.(1) and Figure 1. This is different from RANSAC method in which the outliers have been removed using some theoretical estimated model. (2) The error function for outlier in outlier regularization function is not changed which is different from Huber loss function. In outlier regularization, if the threshold is proper set, the severity of outlier is reduced to that of non-outliers.

### 2.2 Variational representation

The outlier regularization/correction function Eq.(1) is discrete. One important feature of this function is that it also has an equivalent continuous/variational representation. We have the following:

**Proposition 1** $\tilde{y}_i$ *of Eq.(1) is the optimal solution to the following optimization problem:*

$$\tilde{y} = \arg\min_z \|y - z\|_1 + \frac{1}{2\delta}\|z - f\|^2. \quad (2)$$

Proof. Setting $u = z - y$, Eq.(2) can be written as

$$\min_u \delta\|u\|_1 + (1/2)\|u - (f - y)\|^2.$$

The solution of this proximal operator for Lasso is known to be

$$(u^*)_i = \text{sign}(f_i - y_i) \max\left(|f_i - y_i| - \delta, 0\right). \quad (3)$$

Thus $z^* = u^* + y$. If $y_i$ is a non-outlier, $|f_i - y_i| \leq \delta$, $u_i^* = 0$, thus $z_i^* = y_i$ which is the same as Eq.(1). If $y_i$ is an outlier, $|f_i - y_i| > \delta$,

$$\begin{aligned} u_i^* &= \text{sign}(f_i - y_i)(|f_i - y_i| - \delta) \\ &= \text{sign}(f_i - y_i)|f_i - y_i| - \text{sign}(f_i - y_i)\delta \\ &= f_i - y_i - \text{sign}(f_i - y_i)\delta. \end{aligned}$$

Thus $z_i^* = f_i - \text{sign}(f_i - y_i)\delta = f_i + \text{sign}(y_i - f_i)\delta$ which is the same as Eq.(1). □

## 3 Prediction Learning and $L_1$ Norm Function

In our definition of outliers, $f$ is assumed to be known. In some situations, $f$ is known before hands. But in many application, $f$ is not known before hands. We need to learn $f$ simultaneously as we do outlier regularization. This can be done in a successive improvement algorithm as below.

### 3.1 Learning prediction function while outlier-regularization

Here we use linear regression as an example. We set $f_i = a^T x_i + b$, where $(a, b)$ are the slope and intercept, the model parameters to be learned. We cast this simultaneously learning of the prediction $f$ and outlier regularization as outlier-regularized linear regression (ORLR) using the following optimization problem:

$$\min_{\tilde{y}, a, b} \quad \|\tilde{y}^T - (a^T X + b)\|^2, \quad (4)$$

$$s.t. \quad (\tilde{y}_i, f_i = a^T x_i + b) \text{ satisfy } Eq.(1).$$

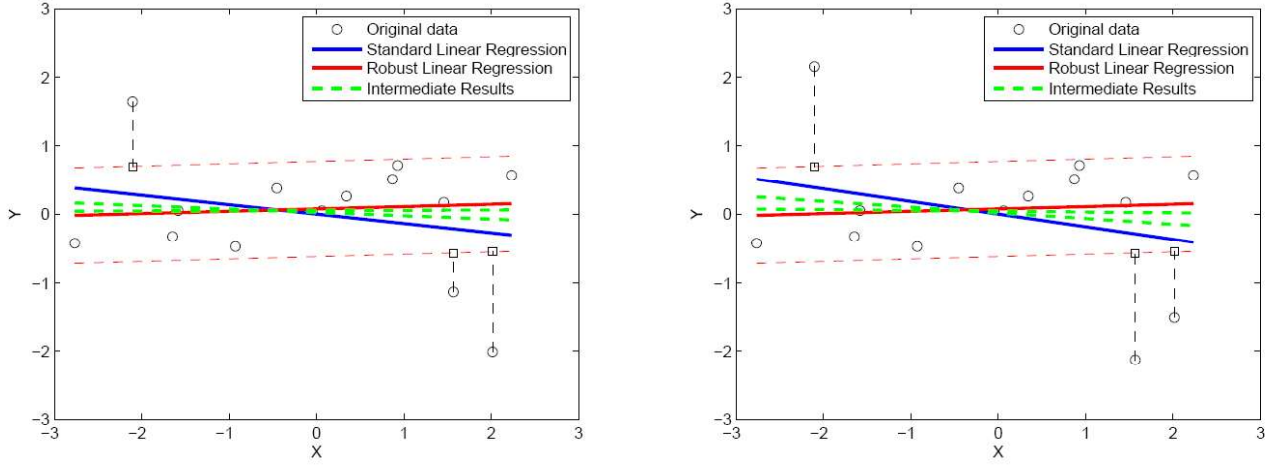

Figure 2: Solution of outlier regularized linear regression. Initial $f$ is the blue line. Second and third improved $f$ are dashed lines. Final converged $f$ is red line. The data in left panel and right panel have same non-outliers, but the three outliers are different. The converged outlier-regularization results are same on both sides, indicating the insensitivity of outlier-regularization w.r.t. the outlyingness of the outliers.

This problem can be solved by the following algorithm:
(A0) Initialize $\tilde{y} = y$.
Repeat Step (A1) and (A2) until convergence:
(A1) Solve the regression problem while fixing $\tilde{y}$.
(A2) Compute regularized $\tilde{y}$ from Eq.(1) while fixing $f$.

In this algorithm, both (Step 1) learning the prediction $f$ and (Step 2) the identification and correction of outliers are successively improved. We note that similar successive update of theoretical prediction and outliers identification have also been used in RANSAC method[Fischler and Bolles, 1981]. The main difference is that the outliers are not only identified but also regularized and corrected in step (A2) of the above algorithm. Figure 2 shows ORLR on a simple dataset with 13 data points including 3 outlier points.

### 3.2 Continuous representation

Using Proposition 1, the prediction function learning step and outlier regularization step in solving Eq.(4) can be combined into a coherent formulation. The new formulation also demonstrates the connection between outlier regularization and $L_1$ function based learning problem.

Using Proposition 1, we substitute $f_i = a^T x_i + b$ into Eq.(2), the optimization problem Eq.(4) becomes

$$\min_{z,a,b} \|y - z\|_1 + (1/2\delta)\|z - (a^T X + b)\|^2. \quad (5)$$

The equivalence of this formulation and Eq.(2) can be seen from the solution algorithm point of view. We can (B0) Initialize $z = y$; (B1) Solve for $(a, b)$ while fixing $z$ and (B2) Solve for $z$ while fixing $(a, b)$. One can see this leads to the same algorithm as in (A0) - (A2) above. Since Eq.(2) is convex, in (B0) $z$ can be initialized to any values.

### 3.3 Small tolerance limit: $L_1$ function

It is noted that the above continuous formulation of outlier regularized linear regression (ORLR) Eq.(5) provides an intuitive understanding of the robustness of $L_1$-norm. Setting the threshold $\delta \to 0$, the regression term is weighted with an infinite weight. Thus $z = a^T X + b$ and the ORLR problem becomes the $L_1$-norm linear regression

$$\min_{a,b} \|y - (a^T X + b)\|_1. \quad (6)$$

This relationship is important because it provides a new explanation on the robustness of $L_1$-norm function, i.e., results of $L_1$ regression is insensitive to how far an outlier lies away from the theoretical prediction line as long as it is an outlier. We will further discuss it in §5.

## 4 Outlier-Regularized PCA

In this section, we apply outlier regularization to PCA. We first extend the outlier regularization function of Eq.(1) to matrix case. We have

$$Z_{ij} = \begin{cases} X_{ij} & \text{if } |X_{ij} - F_{ij}| \leq \delta \\ F_{ij} + \delta \text{sign}(X_{ij} - F_{ij}) & \text{if } |X_{ij} - F_{ij}| > \delta \end{cases} \quad (7)$$

where $\delta > 0$ is a tolerance parameter.

Similar to Proposition 1, the discrete regularization function Eq.(7) on matrix elements also has a continuous representation. We can prove the following Proposition 2 in much the same way we prove Proposition 1.

**Proposition 2** *The regularization function of Eq.(7) is identical to the following problem*

$$Z = \arg\min_{\tilde{X}} \|X - \tilde{X}\|_1 + \frac{1}{2\delta}\|\tilde{X} - F\|_F^2. \quad (8)$$

$\|\cdot\|_1$ denotes the sum of absolute values of matrix entries.

In the following, we apply outlier regularization to PCA to produce a robust formulation of PCA. In PCA, the theoretical prediction is: $F = UV$, where $U \in R^{p \times k}, V \in R^{k \times n}$. Thus, our Outlier-Regularized PCA (ORPCA) can be formulated as,

$$\text{ORPCA:} \quad \min_{Z,U,V} \|Z - UV\|_F^2 \quad (9)$$

$$(Z, X, F = UV) \text{ satisfy Eq.(7)}.$$

ORPCA has several features. (1) It is a robust data representation: the main result of ORPCA is outlier-regularized data $Z$. (2) $Z$ is close to low rank-$k$ subspace; however, $Z$ also contains high rank components. (3) $Z$ can be efficiently computed with computational complexity of $O(kpn)$. We defer detailed analysis of robustness to the next section. Below, we first discuss the continuous formulation of ORPCA and then show that ORPCA becomes $L_1$-PCA at small tolerance limit. At last, we discuss the computational algorithm of ORPCA.

### 4.1 Continuous representation of ORPCA

Using the continuous representation of outlier-regularization function Eq.(8), ORPCA can be equivalently expressed as

$$\text{ORPCA:} \quad \min_{Z,U,V} \|X - Z\|_1 + \frac{1}{2\delta}\|Z - UV\|_F^2. \quad (10)$$

We note that although $U, V$ has exactly rank $k$, $Z$ has much higher rank. More precisely, $Z$ has large components at lower ranks upto rank-$k$, and small components at higher ranks (see §6 in detail).

### 4.2 Small tolerance limit of ORPCA and $L_1$-PCA

We study ORPCA at the small tolerance limit $\delta \to 0$. It is not obvious from the definition of ORPCA of Eq.(9) what will be the limit. However, from the continuous ORPCA representation of Eq.(10), setting tolerance $\delta \to 0$, the 2nd term has an infinite weight. Thus $Z = UV$ and ORPCA becomes a fixed low-rank $L_1$-norm based PCA:

$$\min_{U,V} \|X - UV\|_1. \quad (11)$$

This $L_1$-PCA is a simple and elegant robust formulation of PCA and has been previously investigated [Ke and Kanade, 2005; Baccini et al., 1996; Bolton and Krzanowski, 1999]. The main difference between $L_1$-PCA and ORPCA is that ORPCA retains higher rank components.

The fact that $L_1$-PCA is the small tolerance limit of ORPCA offers some insights into $L_1$-norm property. This process is shown in several steps in Figure 3. At small $\delta$, we are still using the least squares ($L_2$ norm) as the error function as in Eq.(9), but most data points become outliers and are regularized using the regularization function, i.e, pulled towards the theoretical prediction. Clearly, true outliers do not affect the results, and furthermore, the outlyingness of true outliers do not matter either. A deeper understanding of the close relationship between $L_1$ norm and outlier regularization is an important topic to be further investigated.

### 4.3 Computational algorithm for ORPCA

ORPCA can be efficiently computed using the following algorithm.
(C0) Initialize $U = U^0, V = V^0$ and $F = UV$.
Repeat Steps (C1) and (C2) until convergence.
(C1) Fixing $F = UV$. Compute $Z$ using the outlier regularization Eq.(7).
(C2) Fixing $Z$ in Eq.(9), the optimization for $U$ and $V$ is $\min_{U,V} \|Z - UV\|_F^2$. We minimize $U, V$ alternatively. Fixing $V$, we compute

$$U = ZV^T(VV^T)^{-1} \quad (12)$$

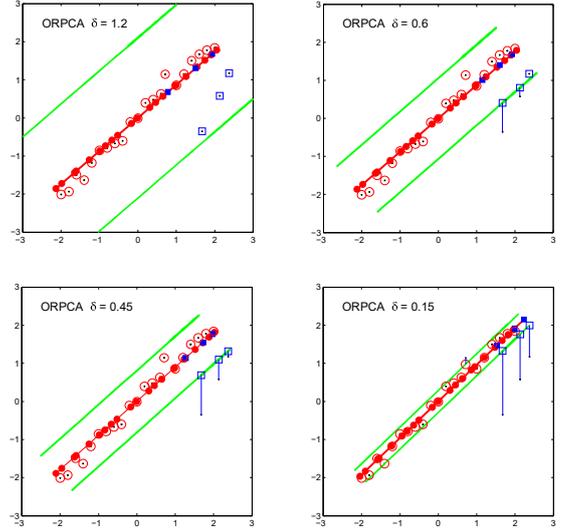

Figure 3: ORPCA on a toy data with 3 outlier. Original data $\{x_i\}$ are shown in black dots. Reconstructed data $\{z_i\}$ are shown as red-circles (non-outliers) and blue-squares(outliers). As the tolerance $\delta$ becomes smaller, the three outliers are pulled closer towards the prediction line.

Note $VV^T$ is a $k$-by-$k$ matrix and its inverse is easily computed because $k \simeq 50$. Fixing $U$, we compute

$$V = (U^TU)^{-1}U^TZ \quad (13)$$

Here, again, $U^TU$ is a $k$-by-$k$ matrix and its inverse is easily computed. The algorithm's computational complexity is mainly computing $U, V$, which is $O(kpn)$. In summary, ORPCA can be efficiently computed. One drawback of ORPCA is its non-convexity. In this paper, we initialize $U^0, V^0$ in step (C0) using the solution of standard PCA.

## 5 Robustness:Insensitivity w.r.t. Outlyingness

One interesting property of outlier regularization function is the **insensitivity** of the results w.r.t. outlyingness of the outliers (how far away they lie from the theoretical prediction): as long as a data point $(x_i, y_i)$ is an outlier, how far away $y_i$ lies does no matter any more, because it is pulled to the tolerance boundary, and thus the final regularization and theoretical prediction/data analysis result remains the same. This can be clearly seen from the discrete formulation of outlier regularization (Eqs.(1,7)). It can also be shown from the following intuitive result.

(1) In Figure 1, we can move the two outliers further away from the prediction function, the outlier-regularization result remains the same: they are on the threshold line.
(2) In Figure 2, the three outliers on the left panel differ from the three outliers on the right panel (the non-outliers are the same). But the results of the outlier regularization (shown as black squares) are the same on both panels.
(3) In Figure 3, the three outliers are pulled towards the prediction. How far away these outliers lie do not matter.

In all these cases, the influence (contribution) of outliers are equal to that of non-outliers. We have the following property:

**Property 1** *Outlier regularization result is insensitive w.r.t. the outlyingness of the outliers. The influence of outliers to the prediction are equal to that of non-outliers.*

Clearly, this property is a form of *robustness*. As discussed, at small tolerance limit, outlier regularization is equivalent to $L_1$-norm function. Thus $L_1$-norm function has the same outlyingness-insensitivity robustness. Therefore, outlier regularization provides a new understanding of $L_1$ robustness from this outlyingness insensitivity point of view. This view differs from the usual understanding of $L_1$ robustness, i.e., in $L_1$ formulation, the residual error (difference between measured data and theoretical prediction) is not squared (as in $L_2$ norm). Thus large residuals due to outliers no longer significantly influence the prediction. This understanding is only partially correct. Although the large residual errors due to outliers are not squared in $L_1$, they are still *large* and thus one would expect they would influence the prediction results *greater* than non-outliers. However, the outlier regularization shows that the influence of regularized outliers are *equal* to that of non-outliers. This type of robustness of outlier outlyingness insensitivity is one main contribution of this work.

## 6 Comparison of ORPCA and RPCA

Outlier regularization is new approach which differs fundamentally from previous approaches in which it (1) provides a clear definition of outlier/corruption and (2) explicitly pulls them back towards the theoretical prediction in concise form. Thus outlier-regularized PCA (ORPCA) has ability to correct gross/large errors in data through outlier-regularization. This is a form of robust data reconstruction which have been studied in many previous works [Torre and Black, 2003; Aanas *et al.*, 2002; Eriksson and van den Hengel, 2010]. Based on the continuous ORPCA formulation of Eq.(10), our ORPCA is close to Robust PCA (RPCA) [Wright *et al.*, 2009] which is

$$\min_Z \|X - Z\|_1 + \beta \|Z\|_{tr}. \quad (14)$$

Here, $\beta$ is a positive weighting parameter. The second term is trace norm $\|Z\|_{tr}$ guarantees that the matrix $Z$ is low rank reduction[Fazel, 2002; Recht *et al.*, 2010]. RPCA has shown good results in recovering underlying low-rank structure $Z$ from the data $X$. A similar work [Liu *et al.*, 2010] is using $L_{21}$ norm. Below, we provide a detailed comparison between ORPCA (Eq.(9) or Eq.(10)) and RPCA.

### 6.1 Computational speed

ORPCA algorithm in §4.3 is easy to implement and fast. It does not involve expansive SVD/eigenvector computation. RPCA uses augmented Lagrangian method [Peng *et al.*, 2010; Lin *et al.*, 2010] and it requires repeated SVD computation. We implemented both methods in MTALAB on a 2.5GHz Pentium computer and compared the running time. The timing results for converging to same tolerance $10^{-10}$ are reported in Table 1. The sizes of different datasets are given in Experiments. The regularized parameter $\delta$ in ORPCA was set to make $Z$ have similar rank with RPCA. Here, we can note that ORPCA is significantly faster than RPCA.

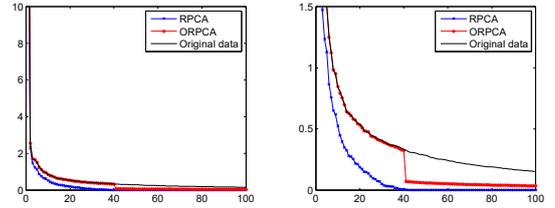

Figure 5: Left: singular values of reconstructed data ($Z$) on AT&T face data. Middle: Vertical axis enlarged singular values. Right: Noise-free residuals from ORPCA and RPCA on AT&T face data. Each class refers to the 10 images of a person.

Table 1: Time (sec) for solving ORPCA and RPCA on five different datasets

|  | AT&Tocc | USPS | MNIST | BinAlpha | COIL |
|---|---|---|---|---|---|
| RPCA | 58.02 | 17.14 | 668.18 | 72.52 | 69.08 |
| ORPCA | 7.78 | 2.98 | 23.79 | 7.25 | 8.14 |

### 6.2 Image reconstruction

To help illustrate the main points, we run both ORPCA and RPCA on the occluded images from AT&T face dataset (more details are given in §7). The original face images are corrupted by gross errors (large square occlusions). Figure 4 shows data reconstruction from RPCA and ORPCA. Due to space limit, we show only two persons with 20 images. Here we observe that (1) Both ORPCA and RPCA reconstruction are robust w.r.t. large occlusion errors. (2) Finer details of individual images are mostly suppressed in RPCA, but are partially retained in ORPCA.

### 6.3 Higher and lower rank components

Figure 5 (left) shows the singular values of reconstructed data $Z$ for ORPCA and RPCA, respectively. We can note that:

First, the singular values of RPCA are substantially suppressed, i.e., almost a constant downshift of the original singular values of the input data. Thus, the important lower ranks $1 \leq k \leq K$ ($K \approx 40$ in this example) are partly suppressed. This can also be seen from the shrinkage result shown in Figure 6. However, in ORPCA model of Eq.(10), the lower rank singular values of ORPCA nearly remain identical to that of input data, i.e., the important lower ranks $1 \leq k \leq K$ are not suppressed.

Second, due to the downshift of singular values in RPCA, higher rank terms ($k > K$) are completed suppressed. In contrast, in ORPCA higher rank components do not appear directly in cost function. They are suppressed, but not completely eliminated. Small but non-zero higher ranks help retain certain fine details in reconstructed images. It can be seen from Figure 4, where ORPCA retains some fine details.

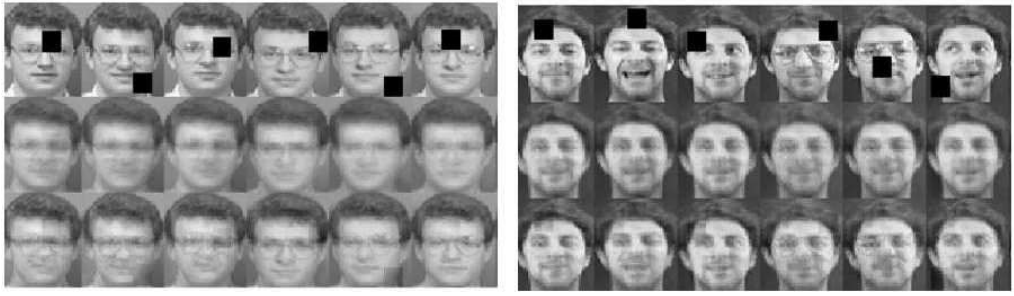

Figure 4: Reconstructions of ORPCA and RPCA on AT&T face data. Top line: original occluded images; middle line: reconstruction from RPCA; bottom line: reconstruction from ORPCA. Finer details of individual images are suppressed in RPCA, but partially retained in ORPCA.

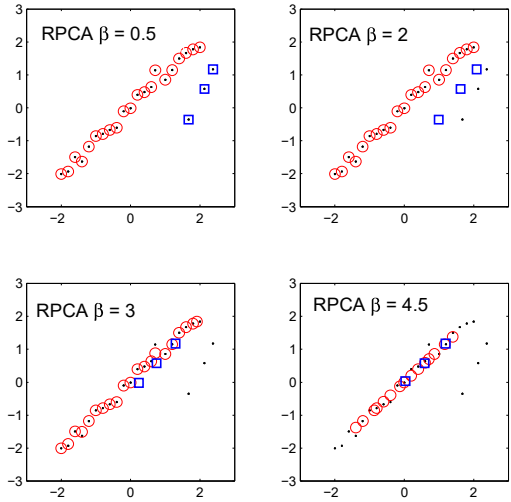

Figure 6: RPCA on the same data in Fig.3. As $\beta$ of Eq.(15) increases, outliers move towards the subspace. But the normal data points shrink due to singular value suppression.

Table 2: Classification results on seven data representations

|          | AT&Tocc | USPS  | MNIST | BinAlpha | COIL  |
|----------|---------|-------|-------|----------|-------|
| Original | 0.825   | 0.551 | 0.585 | 0.511    | 0.846 |
| PCA      | 0.690   | 0.503 | 0.655 | 0.788    | 0.878 |
| L2TrN-PCA| 0.805   | 0.549 | 0.630 | 0.762    | 0.832 |
| RPCA     | 0.887   | 0.611 | 0.690 | 0.551    | 0.905 |
| L1-PCA   | 0.864   | 0.606 | 0.651 | 0.809    | 0.850 |
| ORPCA(Z) | 0.893   | 0.637 | 0.725 | 0.559    | 0.917 |
| ORPCA(V) | 0.908   | 0.614 | 0.715 | 0.816    | 0.916 |

Table 3: Clustering results on seven data representations

|          | AT&Tocc | USPS  | MNIST | BinAlpha | COIL  |
|----------|---------|-------|-------|----------|-------|
| Original | 0.647   | 0.488 | 0.495 | 0.638    | 0.512 |
| PCA      | 0.614   | 0.494 | 0.523 | 0.659    | 0.532 |
| L2TrN-PCA| 0.647   | 0.513 | 0.528 | 0.656    | 0.525 |
| RPCA     | 0.695   | 0.532 | 0.565 | 0.658    | 0.608 |
| L1-PCA   | 0.671   | 0.520 | 0.511 | 0.665    | 0.580 |
| ORPCA(Z) | 0.715   | 0.542 | 0.567 | 0.666    | 0.621 |
| ORPCA(V) | 0.724   | 0.545 | 0.575 | 0.679    | 0.638 |

In general, data should lie in a low-dimensional manifold (low-rank subspace). But this is not absolute. Data should be allowed to have small higher rank components in order to retain certain details. ORPCA is one example of this "mostly low-rank + small higher rank components" model.

## 7 Experiments

We tested ORPCA on several image datasets, as shown in Table 2. In experiments, the parameter $\delta$ in ORPCA was set to 0.003. Our experience is that ORPCA is not sensitive to this parameter. It can return consistent better performance when $\delta$ varies from 0.002 to 0.005. For each image in AT&T face data, we add one 4×5 corruption. We denote this as AT&Tocc.

We first perform classification task on these datasets, and compare our method with some other data representations including original data, standard PCA, RPCA (Eq.(14)), L2TrN-PCA (Eq.(14) with $L_1$ norm changed to $L_2$ norm), L1-PCA (Eq.(11)), ORPCA (Z) of Eq.(11), ORPCA (V) of Eq.(11). In ORPCA, we can either work directly on $Z$ which is ORPCA (Z) above. We can also work on $V$ which is ORPCA (V) above. We use regression classification for this evaluation. All the experiments are performed using 5-fold cross-validation. In additional to classification, we also perform clustering task using different data representations. Here, we run K-means with random initialization 50 times and compute the average clustering accuracy result. Table 2, 3 summarize results. Here, we can note that that (1) PCA performs poorly on occluded data (AT&T), demonstrating PCA is sensitive to large corruption/occlusion noise. (2) Both $L_1$-PCA and RPCA can obtain better performance than PCA, which indicates the robustness of $L_1$-norm function. (3) Compared with RPCA, ORPCA(Z) obtains slightly better performance on both classification and clustering. It clearly suggests the robustness of the proposed outlier regularization function and thus ORPCA model. (4) ORPCA(V) generally performs better than $L_1$-PCA and other data representations.

## 8 Summary

We present a new outlier regularization framework to understand and analyze the robustness of L1 nor function. A key property of outlier regularization is that how far an outlier lies away from its theoretically predicted value does not affect the final regularization-and-analysis results. This reveals the nature L1-norm robustness. We apply outlier regularization to PCA. Comparing to trace-norm based PCA, outlier regularized PCA does not suffer shrinkage effects, retain small high rank components which help retain fine details of the data, and can be computed more efficiently.

## References

[Aanas *et al.*, 2002] H. Aanas, R. Fisker, K. Astrum, and J.M. Carstensen. Robust factorization. *IEEE. Trans. on PAMI*, 24:1215 – 1225, 2002.

[Baccini *et al.*, 1996] A. Baccini, Ph. Besse, and A.De. Falguerolles. An l1-norm pca and a heuristic approach. *Ordinal and Symbolic Data Analysis*, 1:359–368, 1996.

[Bolton and Krzanowski, 1999] R. J. Bolton and W. J. Krzanowski. A characterization of principal components for projection pursuit. *The American Statistician*, 53:108–109, 1999.

[Chandrasekaran *et al.*, 2009] V. Chandrasekaran, S. Sanghavi, P. A. Parrilo, and A. S. Willsky. Sparse and low-rank matrix decompositions. *15th IFAC Symposium on System Identification*, 2009.

[Ding *et al.*, 2006] C. Ding, D. Zhou, X. He, and H. Zha. R1-pca: rotational invariant l1-norm principal component analysis for robust subspace factorization. In *ICML*, pages 281–288, 2006.

[Duda *et al.*, 2001] R.O. Duda, P.E. Hart, and G.D. Stork, editors. *Pattern Classification (2nd ed)*. Wiley Interscience, New York, 2001.

[Eriksson and van den Hengel, 2010] A. Eriksson and A. van den Hengel. Efficient computation of robust low-rank matrix approximation in presence of missing data using the l1 norm. In *CVPR*, pages 771–778, 2010.

[Fazel, 2002] M. Fazel. Matrix rank minimization with applications. *PhD thesis, Stanford University*, 2002.

[Fischler and Bolles, 1981] M. A. Fischler and R.C. Bolles. Random sample consensus: a paradigm for model fitting with applications to image analysis and automated cartography. *Communications of the ACM*, 24:381–395, 1981.

[Huber, 1964] P. J. Huber. Robust estimation of a location parameter. *The annals of Mathematical Statistics*, 35(1):73–101, 1964.

[Ke and Kanade, 2005] Q. Ke and T. Kanade. Robust l1 norm factorization in the presence of outliers and missing data by alternative convex programming. In *CVPR*, pages 739–746, 2005.

[Kwak, 2008] N. Kwak. Principal component analysis based on l1-norm maximization. *IEEE Trans. on PAMI*, 30(9):1672–1680, 2008.

[Lin *et al.*, 2010] Z. Lin, M. Chen, and Y. Ma. The augmented lagrange multiplier method for exact recovery of corrupted low-rank matrices. *UIUC Technical Report UILU-ENG-09-2214*, 2010.

[Liu *et al.*, 2010] G. Liu, Z. Lin, and Y. Yu. Robust subspace segmentation by low-rank representation. In *ICML*, pages 663–670, 2010.

[Lu *et al.*, 2013] C. Lu, J. Shi, and J. Jia. Online robust dictionary learning. In *CVPR*, pages 415–422, 2013.

[Peng *et al.*, 2010] Y. Peng, A. Ganesh, J. Wright, W. Xu, and Yi Ma. Rasl: Robust alignment by sparse and low rank decomposition for linearly correlated images. In *CVPR*, pages 763–770, 2010.

[Recht *et al.*, 2010] B. Recht, M. Fazel, and P. A. Parrilo. Guaranteed minimum-rank solutions of linear matrix equations via nuclear norm minimization. *SIAM Review*, 52(3):471–501, 2010.

[Torre and Black, 2003] F. D.A. Torre and M. J. Black. A framework for robust subspace learning. *Int'l J. Computer Vision*, pages 117–142, 2003.

[Wagner *et al.*, 2012] A. Wagner, J. Wright, A. Ganesh, Z. Zhou, H. Mobahi, and Y. Ma. Toward a practical face recognition system: Robust alignment and illumination by sparse representation. *IEEE. Trans. on PAMI*, 34:372–286, 2012.

[Wang *et al.*, 2012] N. Wang, T. Yao, J. Wang, and D. Yeung. A probabilistic approach to robust matrix factorization. In *ECCV*, pages 126–139, 2012.

[Wright *et al.*, 2009] J. Wright, A. Ganesh, S. Rao, and Y. Ma. Robust principal component analysis: Exact recovery of corrupted low-rank matrices via convex optimization. In *NIPS*, pages 2080–2088, 2009.